# The Field of Safe Motion: Operationalizing the Field of Safe Travel Using Reachability Analysis

Leif Johnson[1*], Trent Victor[1], Johan Engström[1]

## Abstract

We present the Field of Safe Motion (FSM), a quantitative safety model for determining whether a driver maintains a collision-free escape route, or "out," at any given moment by accounting for that driver's physical capabilities and the foreseeable actions of other road users. The Field of Safe Travel (FST) provides a framework for representing the types of sensory information and actions available to drivers. However, the FST has remained conceptual in nature since its initial publication almost 90 years ago—and a concrete computational operationalization is still lacking. At the same time, reachability analysis provides a quantitative basis for assessing the possible actions available to road users, using interpretable kinematic models, but reachability models have so far remained confined largely to the engineering and robotics literature. Bringing these two approaches together provides for an interpretable, quantitative tool for assessing driving behavior across a wide range of driving scenarios. Beyond being interpretable, our approach relies on a relatively small set of basic assumptions that are easy to enumerate and reason about. Furthermore, an interpretable reachability model paired with kinematic assumptions provides a way to bound uncertainty about road users' reasonably foreseeable future locations. We demonstrate the applicability of the FSM to different driving scenarios and discuss the strengths and weaknesses of the model.

## Introduction

More than 80 years ago, Gibson and Crooks (1938) introduced the Field of Safe Travel (FST), which is still one of the most influential models in traffic psychology and a key starting point for most modern attempts to explain and model adaptive driving behavior (e.g., Summala 2007; Kolekar et al., 2020; Ljung Aust and Engström, 2012). The FST broadly represents "the field of possible paths which the car can take unimpeded" (Gibson and Crooks, 1938, p. 454), hinting at Gibson's influential later development of ecological psychology and, more specifically, *affordances* as opportunities for action (Gibson, 1979). The FST thus offers a conceptual framework for how drivers regulate their behaviors based on the future trajectories perceived as available to them at any moment in time.

[1] Waymo LLC, 1600 Amphitheatre Parkway, Mountain View, California 94043 USA
*Corresponding author: leif@waymo.com

In the FST, *valence* represents the influence of perceived excitatory and inhibitory "forces" on future trajectories. A positive valence is assigned to the destination (e.g., a point that a driver wants to reach) and negative valences are pictured in terms of clearance lines surrounding other objects (see Figure 1). Negative valences may be related to any inhibitory force including the risk (probability and severity) of collision or the risk of being sanctioned due to violating the rules of the road (e.g., running a red light).

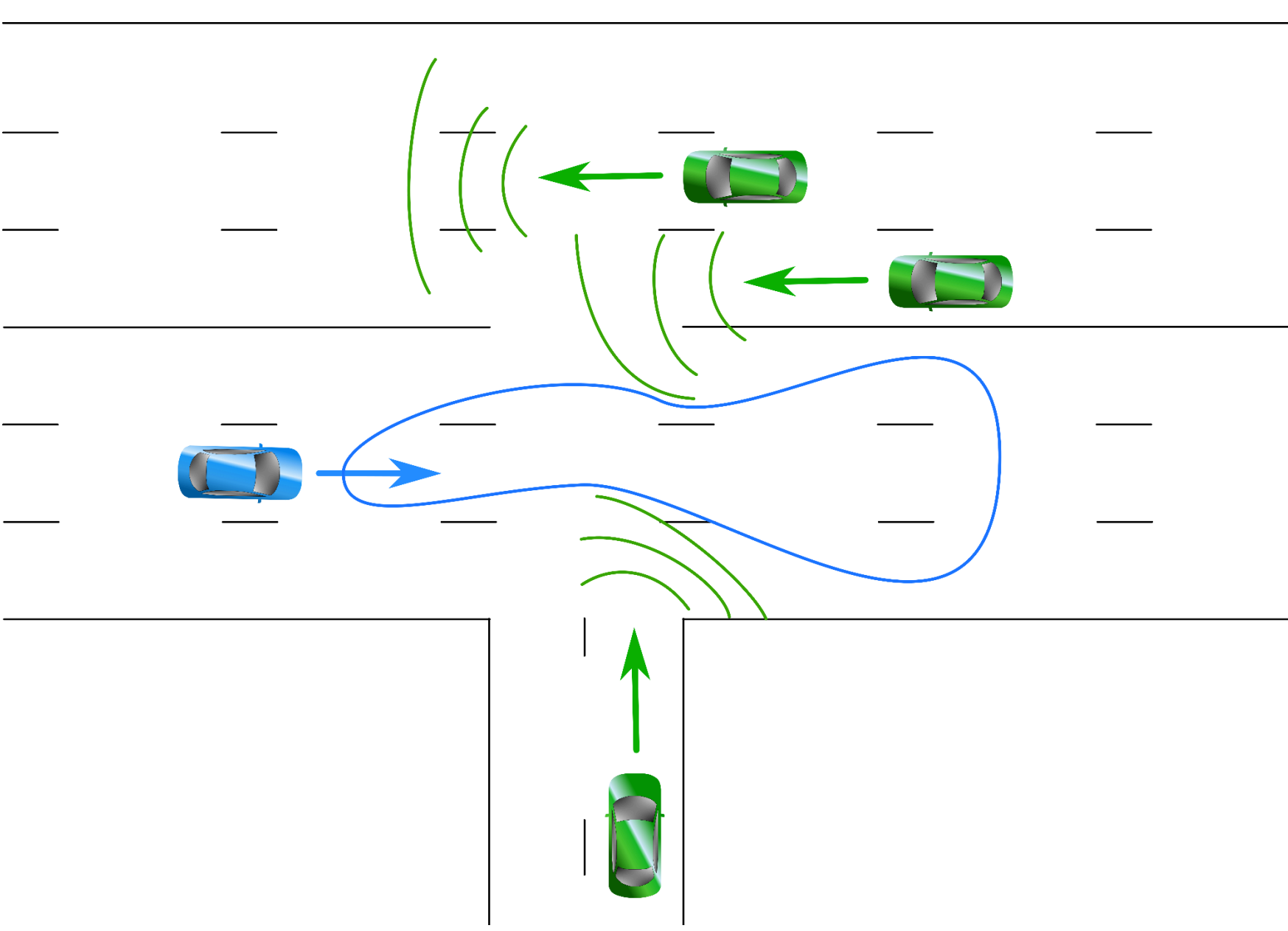

Figure 1. Conceptual illustration based on Gibson and Crook's (1938) Field of Safe Travel. Here, the blue vehicle is the subject vehicle. Green lines illustrate the idea of perceptual "forces" originating from other road users. The blue outline shows the subject vehicle's FST.

In recent decades, the FST has been a main source of inspiration both in ecological psychology (e.g., Kadar and Shaw, 2000; deLucia and Jones, 2017) and traffic psychology, in particular for so-called motivational models of road user behavior (e.g., Wilde, 1982; Summala, 1988, 2007; Fuller, 2005, 2011; Vaa, 2007; see Lewis-Evans, 2012, for a review) and more recent models of adaptive driver behavior (e.g., Ljung Aust and Engström, 2012; Papakostopoulos et al., 2017). Most of these models are focused, like the FST itself, on qualitative descriptions of driving behavior. However, more recent developments in computational human driver modeling use the concept of fields to organize information about the driving scenario. For example, Kolekar et al. (2020) conceives of a "risk field" surrounding the driver and shows how this field can be used to explain aggregate human driving behaviors. Similarly, He et al. (2022) and Mullakkal-Babu et al. (2020) examine probabilistic "risk fields" for assessing risk based on collision probability and severity.

Largely independent of the influence of FST in ecological and traffic psychology, several closely related quantitative safety field models have recently been proposed in the field of vehicle engineering, in particular in the context of automated driving systems (ADS) development and evaluation. These are typically characterized as examples of *safety-related models*, defined in IEEE 2846 (2022) as representations of safety-relevant aspects of driving behavior based on assumptions about reasonably foreseeable behaviors of other road users. For example, Nistér et al. (2019) proposed the Safety Force Field (SFF), which enumerates control policies for all road users that, if universally followed, will avoid collisions. The model incorporates limitations of visibility, attentiveness, and latency and provides a mechanism for layering obstacle avoidance at a low level inside the control mechanisms for an autonomous vehicle. A similar example is Responsibility Sensitive Safety (RSS), proposed by Shalev-Shwartz et al. (2018), which attempts to join low-level kinematic specifications with a high-level interpretation of road user responsibility in different driving situations. RSS assesses driving behavior by introducing the notion of a "proper response" that road users should display in particular situations; for example, the proper response to a decelerating vehicle in front is to brake sufficiently to avoid collision if there is sufficient time and space available to do so. However, RSS does not provide a general mechanism for explaining driving behavior; instead it focuses on collision avoidance in specific driving situations.

In parallel, the related concept of *reachable sets* has gained momentum in the engineering and robotics literature. A reachable set represents the kinematic states (i.e., combinations of position, orientation, speed, etc.) that can be attained at a given point in the future, using Newtonian physics to compute sequences of reachable states over time. Reachability theory offers a powerful computational formalism for representing the driver’s available action space and how those actions might propagate the driver's kinematics in the near future. Although reachable states are in principle an enumeration of all possible future kinematic trajectories, in practice reachability computations are often based on tracking the boundaries of the kinematically reachable set (see Althoff et al., 2021, for a review). Reachable sets have been proposed as a basis for formal requirements and verification criteria for ADS (e.g., Pek et al., 2020; Leung et al., 2020).

In this paper we combine these existing strands of research in psychology and engineering and propose a way in which reachable sets can be used to operationalize the psychologically motivated FST, in a new model we call the Field of Safe Motion (FSM), which differs from the FST in using reachability analysis to operationalize the FST's underlying psychological concepts. We demonstrate how the FSM can be used to evaluate the extent to which human and artificial drivers are able to perceive their available action space and act accordingly. In a sense, the FSM measures whether a driver maintains the possibility of a collision-free escape route (an "out") at any given moment. Specifically, we focus on how the FSM can be used to define precise criteria for both defensive driving and collision avoidance, and evaluate the extent to which human drivers fulfill these criteria.

The key contribution of this paper is the proposal of the FSM as a general quantitative model for understanding and operationalizing safety margins in driving, in a way that generalizes easily to many different types of scenarios while remaining rooted in human psychology. In particular, the FSM is able to incorporate uncertainty in both the driver's own action capabilities, as well as in the behaviors of other road users, where the uncertainty is conditioned on explicit assumptions about the reasonably foreseeable future behaviors of other road users—in line with the notion of a safety related model from IEEE 2846. This model can be used to assess human driving as well as to serve as the basis for driving behavior reference models in the evaluation of behavior of automated driving systems or other Advanced Driver Assistance Systems.

## Model

The Field of Safe Motion measures the drivable space for one road user (henceforth referred to as the *ego*) in a driving scenario. Inspired by the FST, the drivable space for the ego combines the ego's physical capabilities with an estimate of the possible near-future states of the environment, using assumptions about the physical capabilities of all road users. We base the FSM on *reachability analysis* because it provides an interpretable, quantitative method for computing occupied and drivable areas (e.g., Althoff et al., 2021).

In particular, the FSM uses reachability analysis to first compute the limits of each non-ego road user's potentially occupied space at each moment in the near future; each of these limits are called *occupied areas*. After computing the occupied areas for non-ego road users up to a fixed future time horizon, the FSM again uses reachability analysis to compute the ego driver's reachable space. The ego driver's space does not include reachable states for the ego that overlap with any areas that could potentially be occupied by other road users (ORUs). The resulting space for the ego is called the *drivable area* and is used as the basis for computing quantitative metrics (see Figure 2).

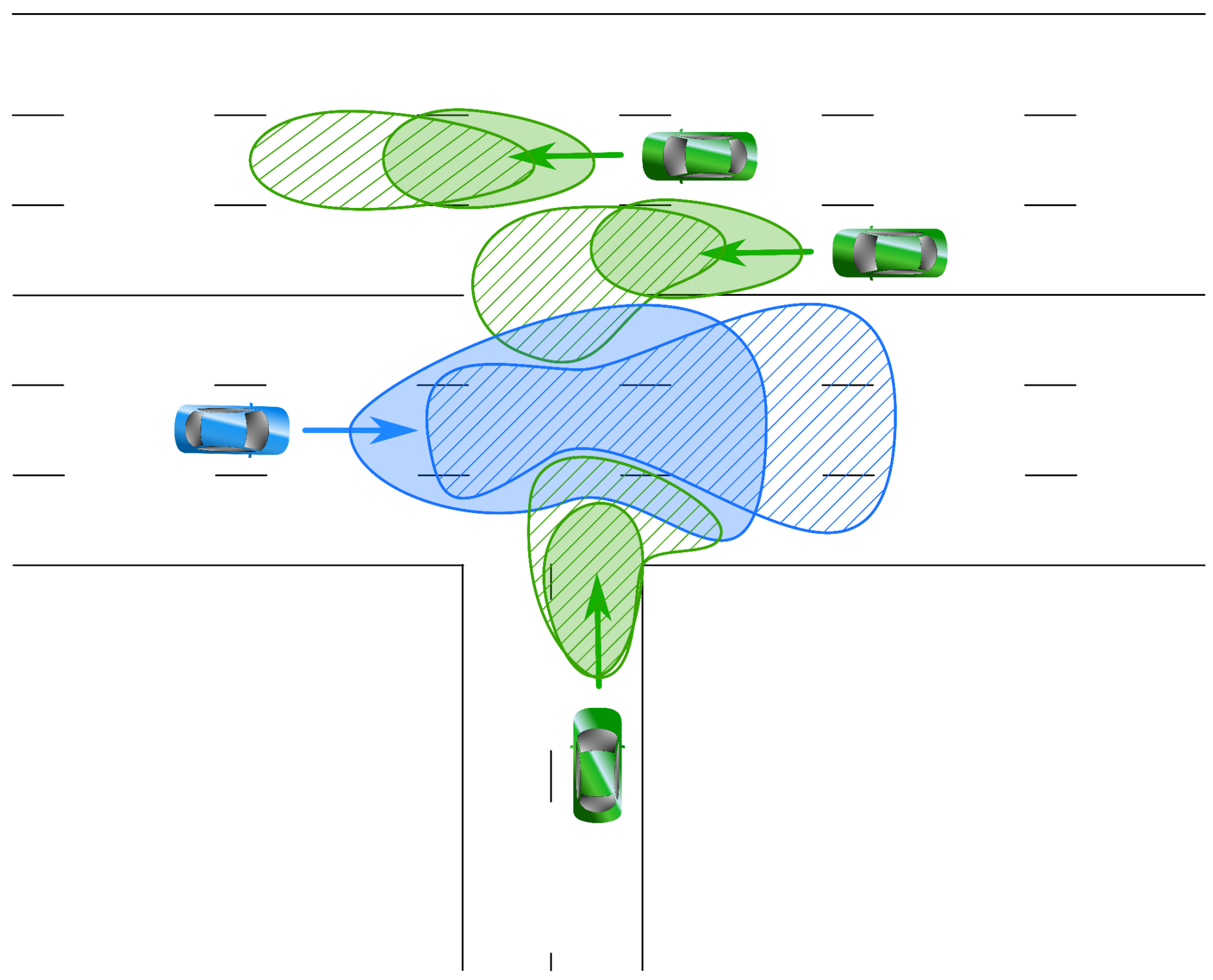

Figure 2. Conceptual illustration of occupied areas for non-ego vehicles (green) and the drivable area for the ego vehicle (blue) at two different times in the future. Solid areas represent reachable sets at future time $\tau_1$ and striped areas represent reachable sets at future time $\tau_2 > \tau_1$.

In addition to standard reachability using kinematic models, the FSM incorporates assumptions based on normative expectations (Bicchieri, 2016; Fraade-Blanar et al., 2026) about socially accepted behavior of other road users, as a way to reduce overly pessimistic estimates of pure reachability-based behavior predictions. Furthermore, the FSM reconceptualizes the standard notion of response time to model the ego's capabilities when faced with surprising future world states. Each of these components is described in more detail below.

## Reachability analysis

Briefly, reachability analysis for a road user computes a sequence of *reachable sets* representing the possible future states of the world, under a set of assumptions governing the foreseeable behaviors of entities in the world in the near future. Each reachable set $R_t(\tau)$ contains the kinematic states that are possible to attain at future time $\tau$, starting from the state of the world at real time $t$. $R_t(0)$ contains just the kinematic state (e.g., location, speed, heading, etc.) of the road user at time $t$. Subsequent reachable sets are, in principle, formed by recursively applying all available kinematic controls to each reachable kinematic state from the previous time step. That is, $R_t(\tau + \Delta\tau)$ contains all kinematic states from $R_t(\tau)$, projected through a kinematic model having time gradient $f'$, using the available controls $U$ at $\tau$, for a discrete time step of $\Delta\tau$. Written symbolically, we get

$$R_t(\tau + \Delta\tau) = \{ s + \Delta\tau \cdot f'(s, u) : s \in R_t(\tau), u \in U(\tau) \}$$

where $s$ and $u$ are elements from the reachable-state and control-value sets, respectively. The reachability computation halts at a fixed time horizon, $\tau = H$.

Conceptually, reachable sets are an exhaustive enumeration of all possible future sequences of kinematic states for an agent. In practice, a complete enumeration is not computationally feasible, so different methods have been developed that trade off computational tractability for a more limited representation of the set of reachable states (see Althoff et al., 2021, for a recent survey). For example, many reachability methods opt to keep track of the states only near the boundaries of each reachable set, since points on the interior of the set are often of less significance than points at the outer edge. That is, the outer edge of the reachable set tends to incorporate kinematic states having, for example, the maximum speed or spatial reach that a road user could achieve, and those extreme states are often of greatest interest when computing things like collision severity estimates or safety margins. Many methods that focus on set boundaries further assume that the reachable set is convex, permitting the sets to be represented efficiently using convex hulls or zonotopes. Even convex sets become expensive to compute, however, as the size of the set, and the dimensionality of the state space, increases. To address this problem, the FSM combines an explicit representation of a vehicle's friction ellipse with the sampling approach εRANDUP proposed by Lew et al., (2022) to compute each reachability step in a constant time, while retaining theoretical guarantees that the resulting set encompasses all reachable states.

## Kinematic rollouts

The FSM uses a single global coordinate system for the locations and orientations of all agents. The positive $x$ axis points east, the positive $y$ axis points north, and the positive $z$ axis points up, i.e., toward the sky. (At this time the $z$ axis is used only for interacting with the roadgraph, which will be discussed later.) A yaw of 0 points east, along the $x$ axis.

For all road users, the FSM uses a jerk-limited kinematic model; kinematic states consist of the following elements:

- $x$ position of the center of the agent *(x)*,
- $y$ position *(y)*,
- yaw *(ω)*, i.e., counterclockwise rotation around the $z$ axis,
- forward speed *(s)*,
- forward (longitudinal) acceleration *(a)*, and
- lateral acceleration *(b)*.

For pedestrians, the underlying kinematic model is a point mass, while other types of road users (e.g., two- and four-wheeled vehicles) use a kinematic bicycle model (e.g., Matute et al., 2019). The main difference between the two models is that the bicycle model's turning capability is limited by the agent's wheelbase and assumed maximum allowed lateral acceleration, while point mass models are limited by an assumed maximum yaw rate.

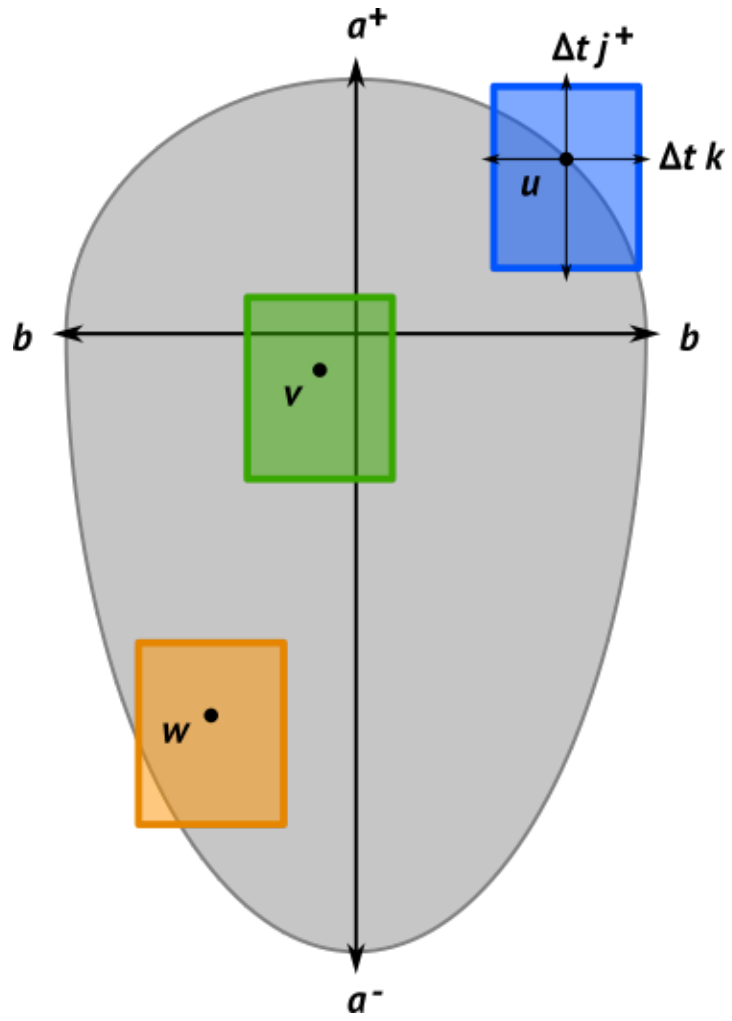

Figure 3. Illustration of the friction ellipse (gray area) for a road user. Points on this diagram are ordered pairs of lateral and longitudinal accelerations, displayed along the horizontal and vertical axes, respectively. Three example points in this space are indicated with letters *u* (blue), *v* (green), and *w* (orange). For each point, a surrounding rectangle indicates the region of acceleration values that are attainable at the next time step, constrained by the jerk limits. However, all valid accelerations must remain inside the ellipse; for example, if an agent is currently accelerating near point *u*, it cannot increase either longitudinal or lateral acceleration without decreasing the other to stay within the ellipse. Point *w* is somewhat constrained in the sense that the agent cannot apply maximal left acceleration and maximal deceleration while staying within the friction ellipse. At point *v*, however, the agent can utilize the full range of its capabilities within the jerk limits.

The FSM limits all kinematic variables except $x$, $y$, and $\omega$ to ranges provided to the FSM at runtime; the assumed bounds of the different kinematic parameters constitute the basic assumptions on which the FSM relies. For example, the speed of an agent is assumed to be limited to the interval $[s^-, s^+]$ such that $s^- < s_\tau < s^+$ at each $\tau$. The acceleration and deceleration of each agent is similarly bounded. In addition to constant limits for speed and acceleration, the FSM also limits the maximum jerk of each agent, which constrains the available ranges of accelerations at each $\tau$, based on the acceleration at the previous $\tau$.

For the longitudinal case, acceleration $a_\tau$ is bounded below and above by separate deceleration and acceleration limits, $a^-$ and $a^+$, and corresponding deceleration and acceleration jerk limits, $j^-$ and $j^+$:

$$max(a^-, a_{\tau-1} - \Delta\tau \cdot j^-) < a_\tau < min(a^+, a_{\tau-1} + \Delta\tau \cdot j^+).$$

The absolute lateral acceleration is bounded by a single maximum, $b$, and associated jerk limit, $k$:

$$|b_\tau| < min(b, b_{\tau-1} + \Delta\tau \cdot k).$$

Finally, the FSM further limits the *combination* of longitudinal and lateral acceleration to lie within the *friction ellipse* for an agent (Brach and Brach, 2011), which ensures that lateral and longitudinal acceleration trade off against one another in a physically realistic way (see Figure 3):

$$(a_\tau / a^+)^2 + (b_\tau / b)^2 < 1 \text{ for } a_\tau > 0$$
$$(a_\tau / a^-)^2 + (b_\tau / b)^2 < 1 \text{ otherwise.}$$

All limits in the kinematic models are applied at each future time step $\tau$, such that any kinematic state with a value exceeding a relevant limit is clipped to the limit before using the state as an input to the equations of motion. (An alternative to clipping is to discard or resample states that go outside the established limits.)

## Normative expectations

Reachability computations relying only on kinematic limits are conservative in the sense that, at the very least, some kinematically reachable states are not physically reasonable (e.g., passing through a physical road barrier), and others should be disallowed for various reasons—including, in particular, conformity to normative expectations such as adhering to the rules of the road.

In the context of road traffic, the notion of *normative expectations* (Bicchieri, 2016), represent the shared beliefs in a society of how one ought to behave on the road (Fraade-Blanar et al., 2025). While some normative expectations (e.g., stopping at a red light) have been explicitly encoded in road rules, normative expectations include any established traffic norms established in a given society. For example, it is generally reasonable to assume that other vehicles will stay in their current lane unless they indicate otherwise: assuming that any road user could occupy any kinematically reachable part of the road would make it difficult to navigate many common roadway scenarios safely. For example, overtaking a vehicle on the freeway is only reasonable under the assumption that the vehicle being overtaken will remain in its lane during the overtake maneuver. As another example, when driving on a two-lane road with one lane in each direction, drivers often pass alongside vehicles traveling in the opposite direction, in their own lane; it is generally reasonable to imagine that each vehicle in such a scenario will remain in its own lane. However, if a road user provides an indication of occupancy of a new lane (e.g., by using a turn signal, by crossing a lane boundary, or by changing yaw enough to make a lane change unavoidable) this new lane can be incorporated into the available lane space for that user.

Thus, to account for normative expectations in addition to assumptions based on kinematic capabilities, the FSM incorporates assumptions based on the roadgraph, which provide critical limits on possible future behaviors of other road users. Many assumptions based on the roadgraph rely on the notion of the *current lane* of a road user. Lanes are assigned to agents in the FSM by considering the footprint of the agent, projected onto the roadgraph. Similar to the process described in Manzinger et al. (2020), the FSM approximates an agent's footprint using the union of a set of circles arranged longitudinally along the centerline of the vehicle.

In addition to the strict footprint of the agent, the determination of lane assignments can also extend an agent's footprint forward to its minimum stopping distance, given its current speed and assumed friction available between the vehicle and the road. Overestimating the footprint of a vehicle based on its observed kinematic state is one way to, for example, take into account vehicles that are about to run a red light, or to extend the reasonably foreseeable lane occupancy of an agent whose yaw is changing enough to cross into a neighboring lane. However, extending footprints forward in this way can also produce erroneous lane assignments when agents are driving in curved sections of road; further work is required to incorporate lateral maneuvers into the minimum stopping distance criterion. For the examples explored in this paper, the minimum stopping distance criterion was not included in agent footprint computations.

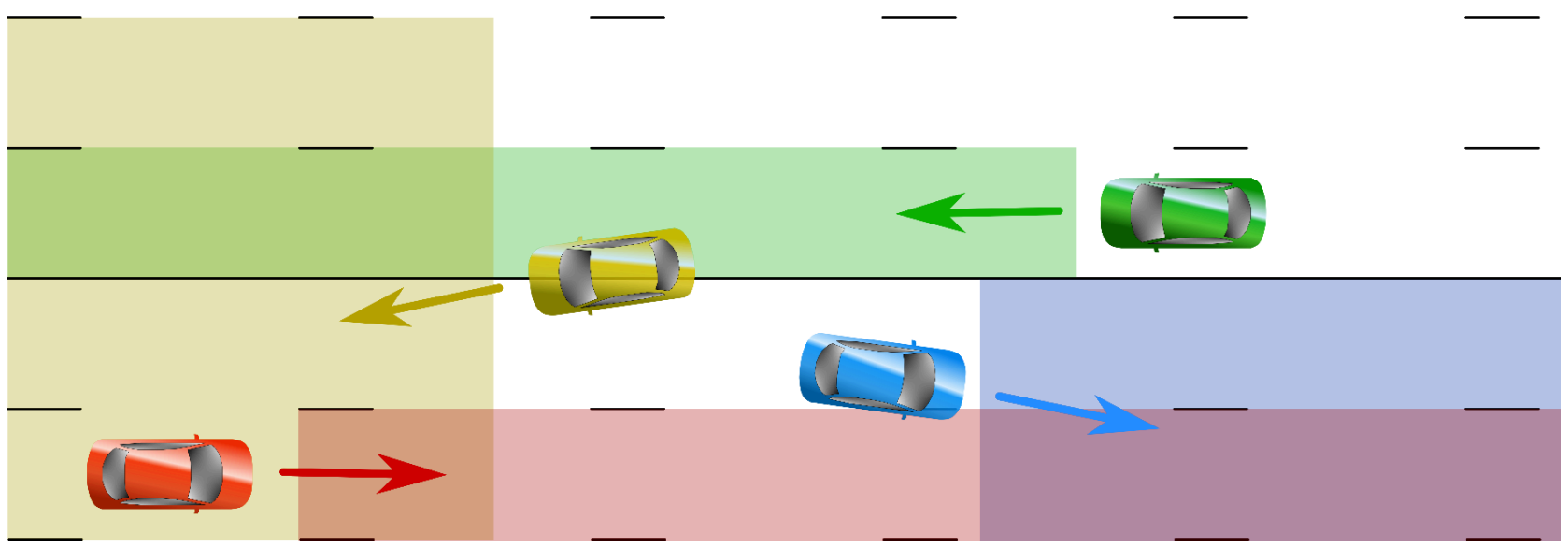

Figure 4. Illustration of lane assignments (shaded rectangles) for various vehicle configurations. The red and green vehicles are assigned a single lane. The blue vehicle is crossing a lane boundary, so it is assigned the two lanes sharing the boundary. The yellow vehicle is anti-aligned with an oncoming lane, so it has no lane restrictions; i.e., its reachable areas are permitted to occupy any lane.

Once the vehicle's footprint has been computed, it is then used to compute a seed set of available lanes for the agent. This seed set contains all lanes whose

- centerlines or boundary lines are covered by the agent footprint,
- directionality lies within $\varepsilon$ of the vehicle's yaw (that is, $|\omega_{agent} - \omega_{lane}| < \varepsilon$), and
- distance in the z direction is less than $k$ meters (to avoid assigning overpass or underpass lanes to agents).

If any available lanes are anti-aligned with the agent's yaw (i.e., $|\omega_{agent} - \omega_{lane}| > \pi - \varepsilon$), then the seed set for the agent is set to the empty set.

Once the seed set is determined, the FSM recursively incorporates all "downstream" lanes from the roadgraph into the full available lane set for an agent (see Figure 4). Finally, the available lane set, if it is not empty, is used to mark areas on the ground plane that are permitted for the agent to occupy. If the available lane set is empty (either because some part of the agent's footprint is anti-aligned with a lane, or because no part of the agent's footprint covers an aligned lane), then the agent is considered not to have any lane-based restrictions on its potential occupancy, and its occupancy is then limited to physical road boundaries.

## Ground plane raster

The FSM follows other approaches in the reachability literature by *projecting* the high-dimensional kinematic state of a road user onto a low-dimensional space before computing state overlaps or other operations that detect interactions between a road user and the surrounding world. In the case of the FSM, we project the kinematic state of a road user—normally consisting of the six dimensions listed above—onto a two-dimensional raster representing the $x$ and $y$ coordinates of the ground plane. For each kinematic state of an agent, the FSM renders a rectangle on the raster that represents the corners of the agent's bounding box. The resolution of the raster, $\delta$, can be configured to trade off between computational complexity and spatial precision, and physical padding can be added to the size of the agent's raster representation to incorporate requirements for spatial margins when determining whether agents have overlapped or interacted.

## Counterfactual surprise and replanning delay

Often, reachability analysis assumes that each agent in a scenario has exactly one associated reachable set at a given point in the future. However, in any scenario with an ego and one or more independent, non-ego agents, there are two important classes of possible futures. In one set of futures, the world unfolds in a way that the ego agent has anticipated and incorporated into its current action plan: in these futures the ego agent does not need to respond to surprising events in its surroundings (c.f., Engström et al., 2024). In the other set of futures, something unexpected happens in the world, and the ego must respond to it.

Traditionally the notion of *response time* is employed to represent a delay when responding to some pre-defined stimulus: for example, in a laboratory setting, a subject might be required to press a button as quickly as possible in response to the sudden illumination of a light. However, as discussed in Engström et al (2024), response time is a difficult concept to map to dynamic stimuli as in real world driving scenarios. To conceptualize and measure response times in the real world, Engström et al. (2024) proposed to define response time as relative to the onset of surprising information that indicates the situation did not play out as expected. This implies that the notion of response time does not apply in a “normal” situation with no surprising stimulus.

In many model-based driving assessment algorithms like RSS (Shalev-Shwartz et al., 2018), response time is realized as a *control delay* $\rho$, such that an agent does not exert any control actions until its response time $\rho$ has elapsed; i.e., $U(\tau) = \varnothing$ for $\tau < \rho$. While this implementation of a response time is fairly reasonable assuming a singular surprising stimulus (e.g., in an experimental setting, the sudden onset of an experimental stimulus, or, in more naturalistic driving settings, a sudden hard brake by an agent in front of the ego), applying the response time to unsurprising situations unduly penalizes the ego agent, effectively freezing the agent during the first $\rho$ seconds of each possible future.

To address this dichotomy of possible futures, the FSM reconceptualizes response time as a *replanning delay*. (See also Schumann et al., 2025 for an agent-based implementation of this concept.) In the set of futures where nothing surprising happens, the ego can behave according to its current action plan, and no control delay is needed to model the ego's reachable states; i.e., $U^{EGO}(\tau) = U_{normal}$ for all $0 < \tau < H$ whenever the reachable sets of ORUs conforms to $U^{ORU}(\tau) = U_{normal}$. In the set of futures where another road user does something surprising, i.e., $U^{ORU}(\tau) = U_{initiator}$, the ego needs time to create a new action plan. During this replanning period, the FSM restricts the ego's available controls to the "normal" range, allowing the ego to continue its current action plan until a new plan is ready; i.e., $U^{EGO}(\tau) = U_{normal}$ for $0 < \tau < \rho$. Generally speaking, once the new plan is available, the FSM attempts to capture the updated plan by allowing the ego to use a larger set of "responder" control actions; i.e., $U^{EGO}(\tau) = U_{responder}$ for $\rho < \tau < H$. Typically the FSM assumes that $U_{normal} \subset U_{responder}$. This mechanism provides a general way of representing the increased reachable space that a new action plan might make available, without committing to the specific details of those action plans.

Even though possible futures can be partitioned into this "normal-normal" and "initiator-responder" dichotomy, the FSM used in the remainder of this paper makes the assumption that the "normal-normal" sets of futures are subsets of the "initiator-responder" sets. To see how this can hold, first observe that the reachable set for a "normal" ORU is constrained to expand at a smaller rate than for an "initiator" ORU, so reachable sets for "normal" ORUs are subsets of reachable sets for "initiator" ORUs. Similarly, reachable sets for the ego vehicle are the same for $0 < \tau < \rho$ in both "normal" and "responder" cases; however, the ego's "normal" reachable sets for $\rho < \tau < H$ are subsets of the "responder" reachable sets. Thus, in practice, the FSM computes the "initiator-responder" reachability sets only, but the distinction between normal and surprising futures remains a useful mental model, and could be a source of more detailed driving performance metrics.

## Algorithm

Now that the necessary model elements are in place, we can summarize the computations performed by the FSM. The FSM computes the drivable area for the ego agent by generally following the reachability procedure outlined in Manzinger et al. (2020). This procedure performs two high-level steps at each real time *t*.

First, for each **non-ego** road user:

1. The FSM computes a reachable set based on ORU kinematic limits. This reachable set is computed using the union of two separate sets of kinematic point samples:

a. The first set consists of $n_{PER}$ kinematic states that are evenly spaced around the perimeter of the friction ellipse. Each of these states represents the rollout of one fixed-control trajectory. For example, a point sample at the maximal negative extent of the friction ellipse represents a trajectory where the agent applies maximal braking, while a point sample along the upper-right edge of the friction ellipse, such as point $u$ in Figure 3, represents a trajectory with a fixed combination of acceleration and steering to the right.

b. The second set consists of $n_{INT}$ kinematic states that are sampled from the interior of the friction ellipse and propagated forward using the εRANDUP algorithm (Lew et al., 2022). These states represent trajectories that apply sub-maximal controls and are useful for maintaining the extent of the reachable set in cases where points from set (a) are pruned.

2. The FSM then projects the union of these kinematic states onto the ground plane raster, and prunes (removes) kinematic states that are out-of-lane or off-road when projected onto the ground plane. Note that interactions among non-ego agents are not considered here, which permits parallel computation.

3. The pruned set of reachable states, projected a final time onto the ground plane raster, is then called the occupied area for that road user.

Then, for the **ego** agent:

1. The FSM computes a reachable set for the ego using the same procedure as described in step (1) above. The ego's reachable set is based on "normal" kinematic limits before the replanning delay during the rollout, and then uses "responder" kinematic limits afterward.

2. The FSM projects and prunes these kinematic states based on the ego's permitted areas of the roadgraph, using the same procedure described in step (2) above.

3. The pruned reachable set for the ego is then further pruned based on the occupied areas of other road users at the corresponding future time. Any kinematic states that overlap with occupied areas of other road users are removed from the ego's reachable set. This second pruning step eliminates ego states that are kinematically reachable but whose projection onto the ground plane raster overlaps with areas that could be occupied by other road users at a specific point in the future.

4. The resulting pruned reachable set for the ego, again projected onto the ground plane raster, is called the *drivable area*. The drivable area consists, by construction, of areas on the ground plane that are (a) possible for the ego to reach, (b) consistent with assumptions about the roadgraph, and (c) free from overlap with the anticipated occupancy of other road users.

Finally, once the drivable area has been computed for all real time steps, the FSM uses these drivable areas to compute quantitative metrics for assessing the ego's driving behavior.

## Metrics derived from reachable sets

Using the procedure described above, the FSM is able to compute an allowed, kinematically feasible, conflict-free drivable area for the ego road user, taking into account replanning delays and responses to potentially surprising counterfactual futures. The FSM computes drivable areas for each future time $\tau$, starting from each real time step $t$, in a driving scenario under analysis. Once drivable areas are computed for all $t$ and $\tau$, the FSM can compute overall performance metrics using the entire collection of drivable areas.

Many possible metrics can be derived from the computed drivable areas, including, but certainly not limited to:

- The existence of a drivable area at a particular time $\tau$ in the future, which measures whether there is an overlap-free kinematic trajectory (an "out") starting at time $t$ and extending to future time $\tau$. Especially notable here is the existence of a drivable area at the future time horizon $H$, which indicates the presence of an "out" extending to the time horizon.
- The size of the drivable area at a particular time $\tau$ in the future, which measures the absolute physical space available for an "out" at $t$. This metric can be thought of as a rough proxy for the number of available "outs" at $t$.
- The change in size of the drivable area at a given point $\tau$ in the future, relative to some previous real time step $t-k$. Such a metric represents aspects of "urgency" in a scenario over the course of $t$.
- The "generalized time-to-collision," defined for each $t$ as the future time $\tau$ when the drivable area for the ego vanishes. In other words, the generalized TTC can be thought of as the ego's longest available rollout. This metric can be thought of as representing the urgency at which a driver needs to take action to avoid ending up in a risky situation. Traditional time-to-collision metrics compute temporal proximity to a single obstacle under simplified assumptions (e.g., constant velocity/acceleration/heading) about the future paths of conflict partners, but the generalized version incorporates all kinematically reachable trajectories by both the ego and other road users.

For the remainder of this paper we focus on the first metric, the existence of a drivable area extending to the time horizon; that is, whether $R_t(H) \neq \varnothing$ for each real time $t$. If no drivable area exists at the time horizon for some time $t$, we say the ego is experiencing an FSM *frame violation* at $t$. Frame violations in the FSM could occur for one of two reasons:

- **Ego-initiated**: the behavior of the ego itself limits its drivable area (e.g., failing to maintain a sufficient following distance); or
- **ORU-initiated**: the behavior of other road users limits the ego's drivable area (e.g., a lane change in front of the ego).

The second type of violation will never be completely unavoidable, because the ego does not control the driving behavior of other road users: for example, an adversarial road user could intentionally collide with the ego vehicle. Furthermore, differentiating between ego- and ORU-initiated FSM violations is difficult in general, because one could argue that the ego could drive with a large enough safety margin to protect against many types of "typically unexpected" behavior on the part of other road users. However, such precautions would come with additional costs, such as overly conservative driving behavior.

Having defined a generally applicable FSM metric for each time $t$, we further define (a) the *frame violation rate* as the fraction of times $t$ that have a frame violation, and (b) an overall *FSM violation* to be $r$ continuous seconds of frame violations. For example, if an overall FSM violation is defined as one second of continuous frame violations, then the ego is driving in such a way as to not have an "out" for one full second of real time. The $r$ parameter provides a sort of sensitivity/specificity tradeoff: small values of $r$ will result in more frequent, but perhaps irrelevant, FSM violations, and large values of $r$ will result in less frequent violations, but the violations that are captured will probably be more notable.

## Assumptions

We now summarize the assumptions that the FSM uses to compute reachable futures in driving scenarios. These assumptions are ultimately based on societal consensus, representing reasonably foreseeable expectations on ORU and ego behavior that stakeholders can agree upon (see IEEE, 2022). Assumptions are divided into kinematic values (Table 1) and non-kinematic values (Table 2). Generally speaking, kinematic assumptions govern the kinematic reachability of road users, while non-kinematic assumptions govern the overall behavior of the model. Non-kinematic assumptions typically represent computational tradeoffs between fidelity and speed.

Table 1. Kinematic assumptions for light vehicles. Limits for "normal" ego acceleration and jerk were set to the 90th percentile of observed light vehicle kinematics in a large sample of human drivers observed by Waymo vehicles. Limits for "surprising" behavior, which in this paper apply to "initiator" ORUs as well as to "responder" ego behavior, were set to observed 99.99th percentiles.

| Parameter | | Normal | Surprising |
|---|---|---|---|
| Replanning delay | $\rho$ | 1000 ms | N/A |
| Speed | $s^- .. s^+$ | 0 .. 40 m/s | 0 .. 40 m/s |
| Long. acceleration | $a^- .. a^+$ | -2.2 .. 2.5 m/s$^2$ | -7.3 .. 6.9 m/s$^2$ |
| Lateral acceleration | $b$ | 1.8 m/s$^2$ | 6.3 m/s$^2$ |
| Longitudinal jerk | $j^- .. j^+$ | -1.8 .. 1.3 m/s$^3$ | -6.0 .. 5.3 m/s$^3$ |
| Lateral jerk | $k$ | 0.8 m/s$^3$ | 4.5 m/s$^3$ |

Table 2. Non-kinematic assumptions for the FSM.

| Parameter | | Value |
|---|---|---|
| Time step | $\Delta\tau$ | 100 ms |
| Future time horizon | $H$ | 4000 ms |
| Lane alignment threshold | $\varepsilon$ | 30° |
| Ground plane raster resolution | $\delta$ | 30 cm |
| Friction ellipse interior samples | $S_{INT}$ | 100 |
| Friction ellipse perimeter samples | $S_{PER}$ | 32 |

Kinematic assumptions in the FSM can be provided separately for different types of road users (e.g., cars, pedestrians, bicycles, etc.). In theory, kinematic assumptions could even be applied to each road user individually—this could, for example, allow the FSM to take into account agents that behave nominally compared with agents that are not displaying nominal behavior (e.g., swerving erratically, stopping and starting frequently, etc.). However, the FSM as described in this paper only relies on assumptions about types of road users (e.g., light vehicles, pedestrians, cyclists, etc.). For the examples that we examine below, the only types of road users present are light vehicles.

## Case studies

To illustrate the behavior of the FSM in practice, we applied the model to examples of human driving behavior that Waymo vehicles have observed in the course of their normal operations.[2] These examples demonstrate that the model applies to many different types of driving scenarios, while also permitting exploration of the role that model assumptions play in metric outcomes.

### Following distance

Some human drivers follow vehicles in front of them with an insufficient safety margin (i.e., "tailgating"), while others follow lead vehicles with a larger following distance. We applied the FSM to two different human drivers observed simultaneously on a freeway, to explore the comparative safety margins of these human drivers' following distances. In this scenario, shown schematically in Figure 5, the two human drivers of interest are in the same lane, following a lead ORU. One human driver, referred to as the *tailgater*, follows the ORU in front of it at a close distance (approximately 0.5 s time gap; red vehicle in Figure 5), while a second human driver, referred to as the *follower*, follows the tailgater with a larger distance (approximately 1.5 s time gap; blue vehicle in Figure 5). The FSM reveals that the tailgater does not maintain a consistent drivable area, thus exposing it to surprising actions and possibly putting the driver at risk of a conflict or a collision. Note that the FSM does not require an interaction to result in a collision—or even a conflict—to identify its lapse of drivable area.

This example provides a concrete way to explore the role that assumptions play in the FSM. As described above, the FSM incorporates assumptions about the limits of kinematic parameters, as well as assumptions related to future lane occupancy.

---

[2] Waymo vehicles are autonomous vehicles that are equipped with a set of sensors for perceiving the external driving environment. While driving on the road, Waymo vehicles can record interactions between other vehicles in their vicinity using their sensors.

We will explore the impacts of several changes of assumptions for the ego vehicle that change the FSM's output: in these examples, we use the ego's frame violation rate as a basis for comparison. Drivers that maintain a lower frame violation rate can be said to be driving more safely—that is, maintaining an "out" more consistently—than those with a higher violation rate.

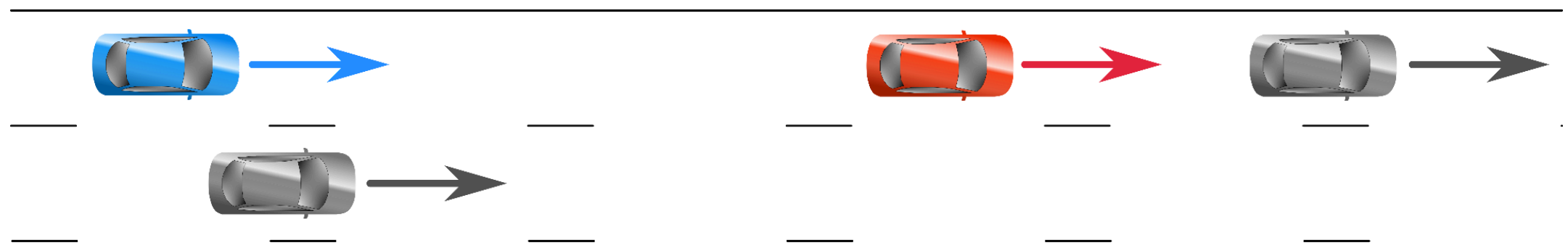

Figure 5. Configuration of vehicles in the following distance scenario. The driver in the blue car follows the vehicle in front of it (the red car) with a large time gap. The driver in the red car follows the car in front of it (the rightmost gray car) with a small time gap.

**Base Case** We compare each change described below with a default set of "base case" assumptions. This set of assumptions is made up of the limit values for the ego and ORU vehicles shown in Table 1. Using these assumptions, the FSM produces drivable area locations for the ego vehicles, which are shown in Figure 6.

The FSM reveals drivable areas visually in Figures 6a (for the follower) and 6b (for the tailgater). In the follower's case, the ego has a drivable area at the time horizon that spans the ego's current lane as well as the areas to the left and right. Because the ego has a nonempty drivable area at the time horizon, there must exist an "out"—a sequence of conflict-free, nonempty reachable sets for the ego extending from $\tau = 0$ to $\tau = H$. Because the follower (Figure 6a) has drivable areas in its current lane as well as neighboring lanes, it must have "outs" that involve braking only (ending up in the current lane), as well as "outs" that involve some amount of steering to the left or to the right. In contrast, the tailgater (Figure 6b) only has a drivable area in the shoulder on the left, indicating that it is following the vehicle in front of it so closely that there is no "out" involving only braking, and the trajectories for the tailgater that involve steering to the right could collide with ORUs in other lanes.

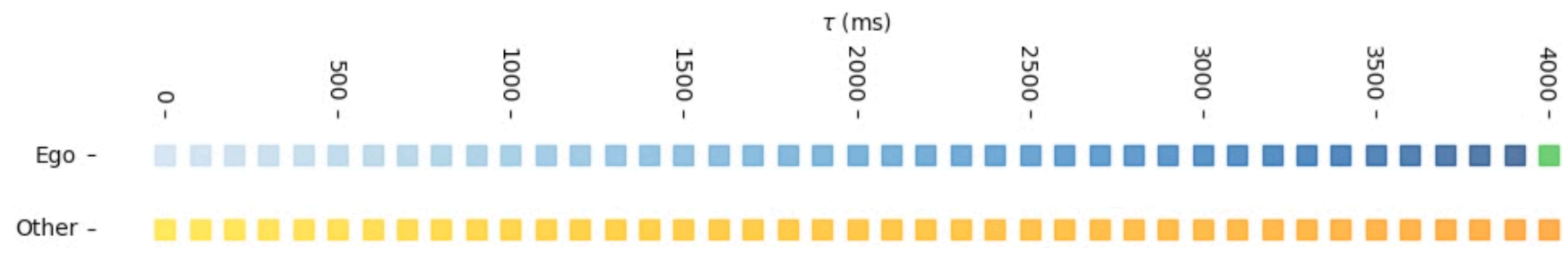

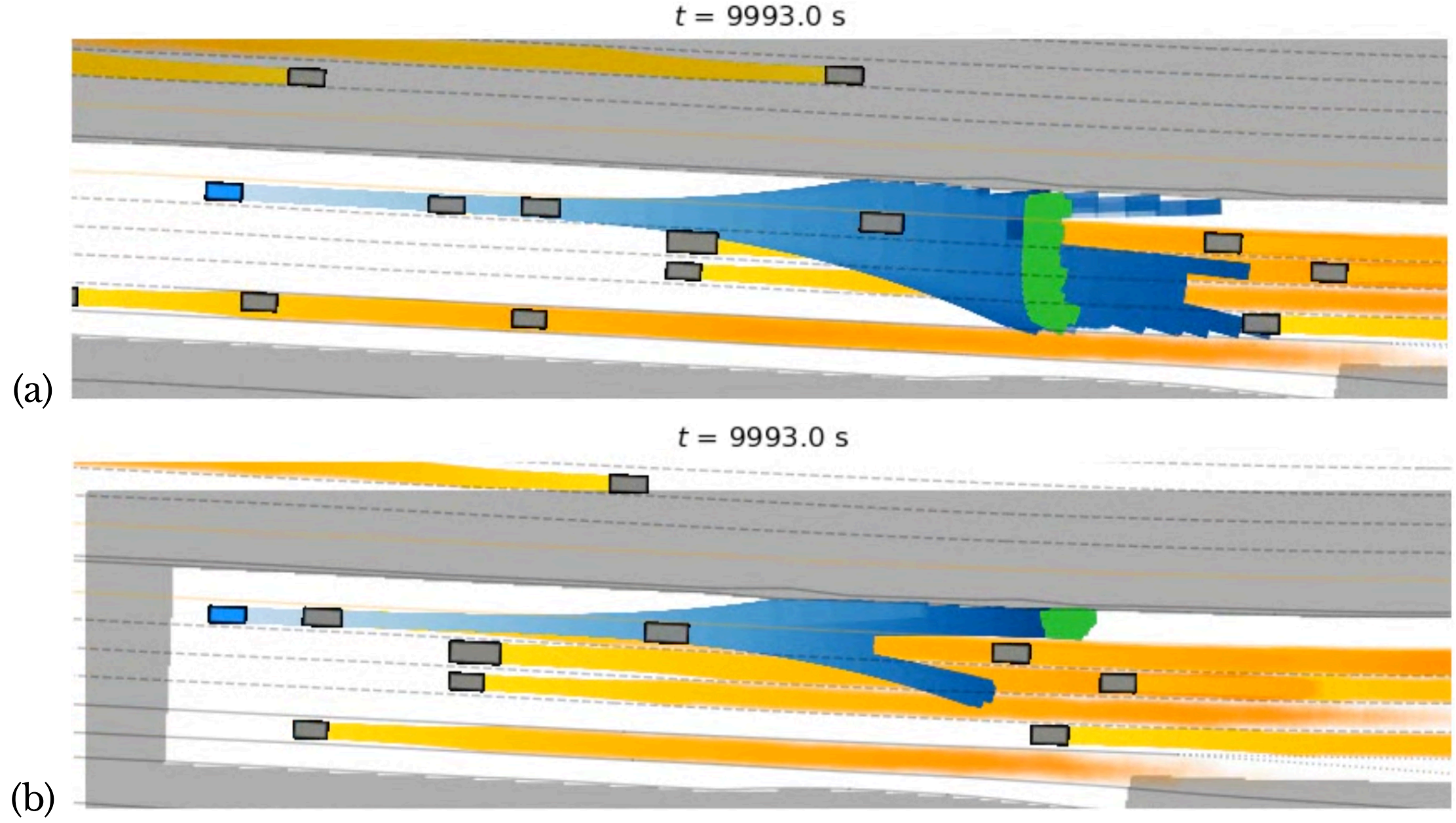

Figure 6. Visualization of reachable sets for the base case, from the perspective of the follower vehicle (panel a) and tailgater vehicle (panel b). Prohibited areas of the road graph are shown in gray, while permitted areas of the road graph are shown in white. Initial positions of road users are shown using outlined rectangles, with reachable sets shown as colored polygons. Reachability polygons for the ego transition from lighter to darker blue as future time $\tau$ increases from *0* to *H*, and reachability polygons for ORUs transition from lighter to darker orange as future time increases. Green areas show the ego's drivable area at the future time horizon, i.e., the set of conflict-free positions on the road that the ego can kinematically occupy at $\tau = H$. Panel (b) shows that the tailgater vehicle only has a drivable area at the *4000 ms* time horizon that involves steering to the left, onto the shoulder of the road.

**Replanning Delay** The base replanning delay value represents an assumption that the ego requires 1000 ms to compute and apply a new action plan in response to a surprising ORU action. As discussed above, the FSM uses the larger set of responder kinematic parameters to represent potential actions after formulating a new action plan. Thus, when performing future rollouts, the ego uses the "normal" kinematic parameters for $\tau$ < 1000 ms; subsequently the ego uses "responder" kinematic parameters. We held the other base case assumptions constant while altering the replanning delay for the ego, to measure the effect that this parameter has on the model results (see Figure 8a).

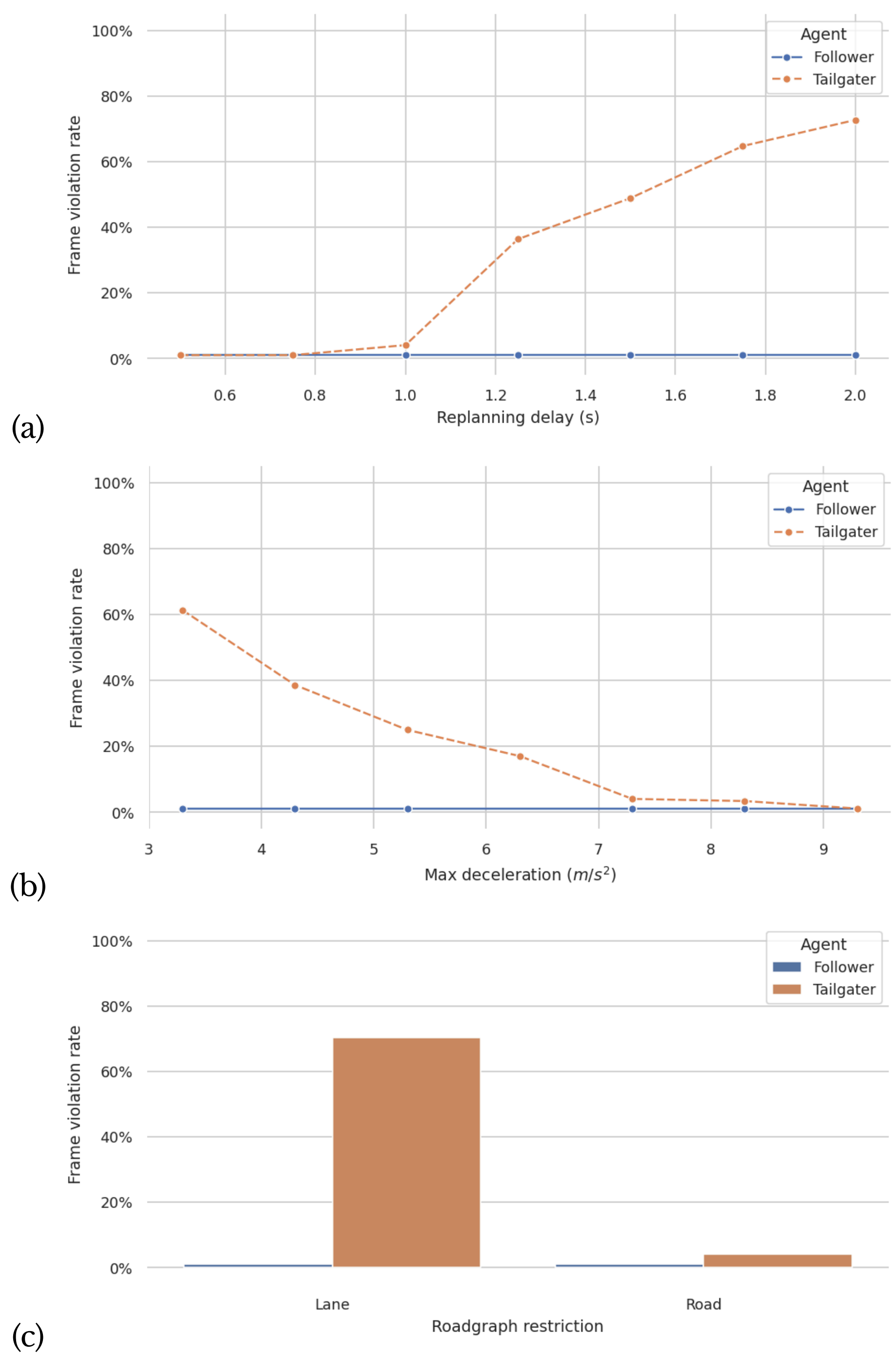

Figure 8. FSM frame violation rate for the two human drivers in Figure 5 as a function of the ego driver's assumed replanning delay (panel a), maximum deceleration (panel b), and assumed roadgraph restriction (panel c).

With a replanning delay greater than 1000 ms, the tailgater has an increasing frame violation rate (vertical axis in Figure 8a), while with a replanning delay less than 1000 ms, the tailgater is able to maintain a low frame violation rate. Since the follower has a larger safety margin, changes in its assumed replanning delay have no effect on the frame violation rate.

**Longitudinal Acceleration** As described above, accelerations in the FSM are limited to a constant range provided to the model. Changing these assumptions results in different patterns of future overlaps between the ego agent and other road users. Here, we varied the deceleration limit for the ego vehicle from the base assumption (see Table 1) and measured the results from the FSM.

As with the replanning delay, assumptions about the braking limit affect the tailgater but have negligible impact on the follower's frame violation rate, as shown by the orange and blue lines in Figure 8b. If the FSM was used to identify unsafe driving behaviors by searching for high frame violation rates, the braking limit parameter would have little effect on the outcome for drivers that maintain larger safety margins, while this parameter does have an effect on drivers like the tailgater who maintain smaller safety margins.

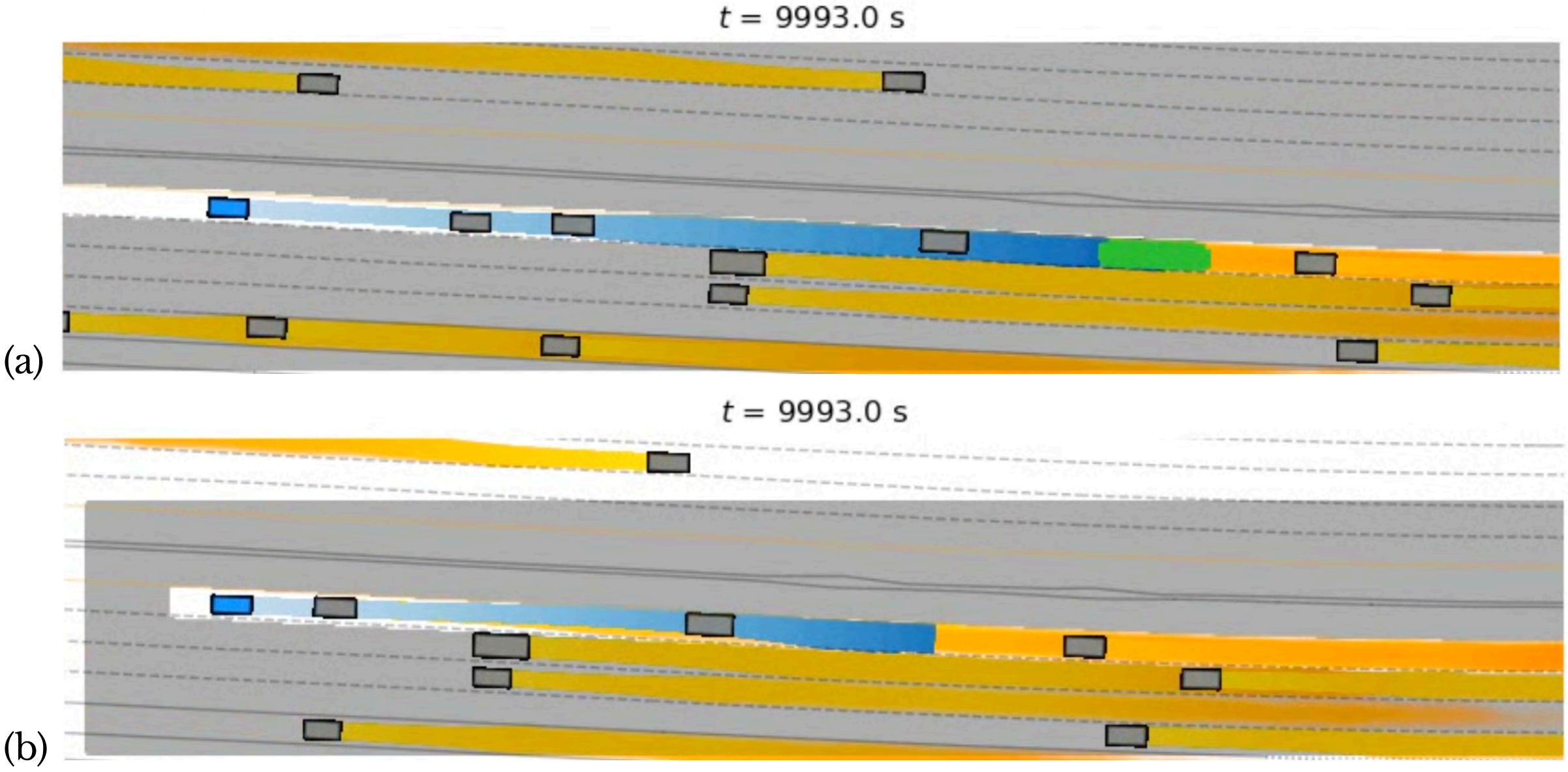

Figure 7. Visualization of reachable sets from the perspective of the follower (panel a) and tailgater (panel b), under an assumption that the ego vehicles will remain in their current lane. See Figure 6 for descriptions of the colors and plot elements.

**Roadgraph Restriction** As shown in Figure 6b, even a driver that maintains an insufficient longitudinal time gap for braking is still able to maintain its drivable area by steering. For comparison, we also computed the FSM using a more restrictive assumption about the roadgraph; namely, that the ego driver will stay in its current lane (see images of the outcomes in Figure 7 and aggregate results in Figure 8c). This restriction had a large effect on the tailgater, as shown by the change in the orange bars in Figure 8c, but no impact on the follower, who has enough distance to the vehicle in front to permit a drivable area while remaining in lane.

## Same-direction lateral incursion (SDLI)

We applied the FSM to an example same-direction lateral incursion (SDLI) event that the Waymo Driver observed between two human-driven vehicles in the course of its normal operations. In this scenario, one vehicle (the ORU initiator) changes lane in front of another (the ego responder) while both vehicles are driving the same direction on a multilane road.

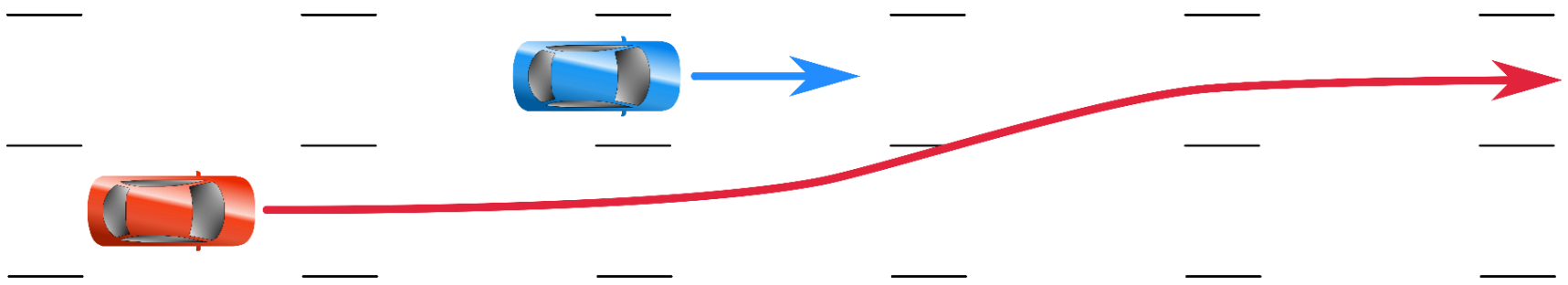

Figure 9. Illustration of key driving interaction for the SDLI scenario. The ego vehicle (blue) gets cut off by the other road user (red) and then the ego vehicle brakes to restore its FSM.

An initiator's lane change can restrict the FSM available to the responder by removing some of the responder's drivable area. In such a case, the responder should initiate a braking or steering response to restore its FSM as quickly as possible. The Waymo vehicle observed a braking in response to a lane change in this SDLI scenario: once the responder's drivable area was eliminated by a lane change, the responder applied the brakes, decelerating until the drivable area reappeared (see Figure 10). Once the drivable area reappeared, the responder accelerated slowly to restore its original speed.

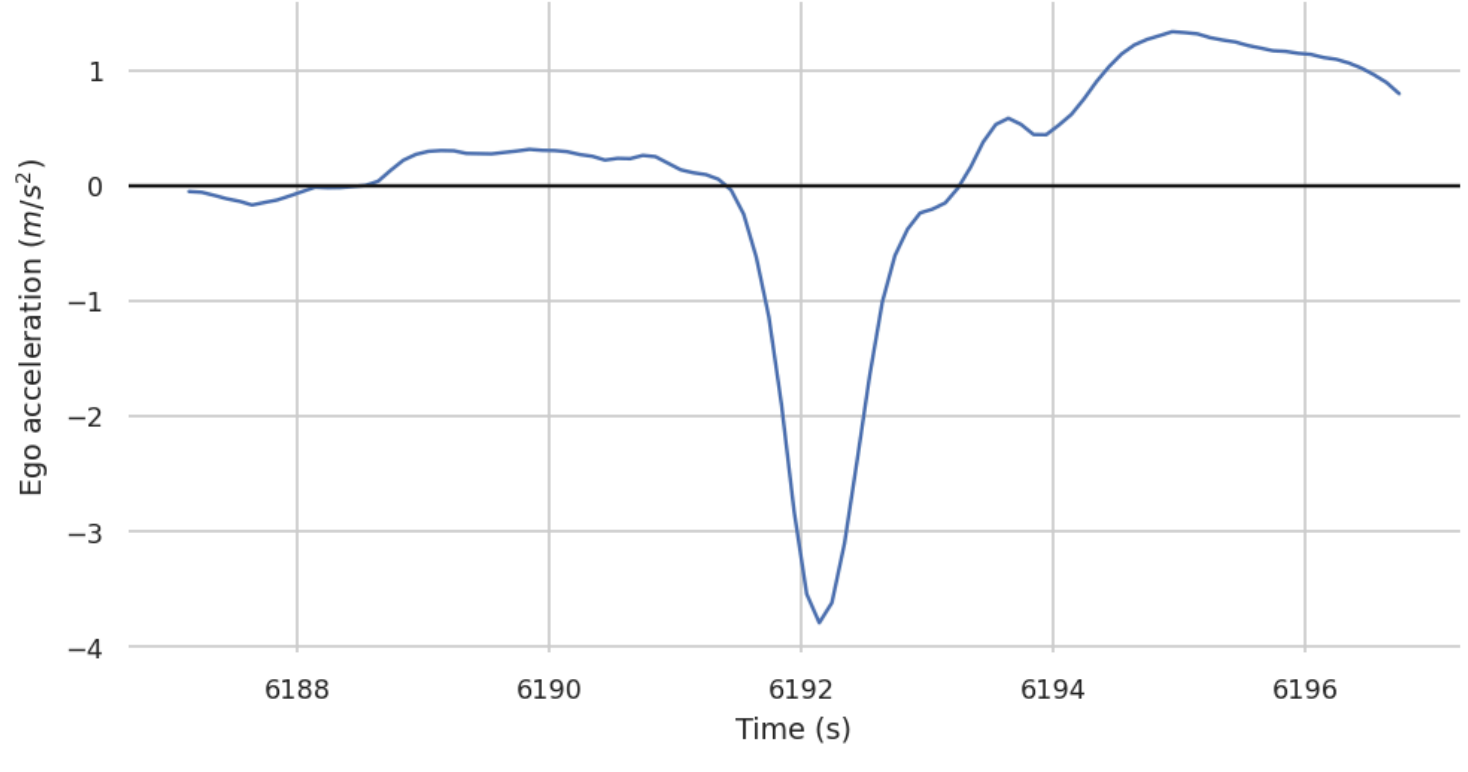

Figure 10. Acceleration for the ego in the SDLI scenario.

The FSM can provide detailed spatiotemporal resolution of these reaction processes at work (see Figures 10 and 11). In this example, at time 6190, prior to the SDLI event, the ego (blue) has a drivable area extending to the future time horizon $H = 4000\ ms$ (green areas near the bottom of Figure 11a). At time 6191.5, the lane incursion is in progress, and the ego brakes (see Figure 10). At this time, the ego does not have a drivable area extending to the future time horizon (no green areas in Figure 11b). At time 6193, the ego's braking behavior has reduced its speed and restored a drivable area at the future time horizon (green areas near the bottom of Figure 11c).

Finally, it is worth noting here that the assumptions used for the ego and ORUs (including the initiator of the lane incursion) were the same as those used for the "base case" in the previous scenario. No additional tuning of kinematic parameters was necessary to analyze the SDLI event.

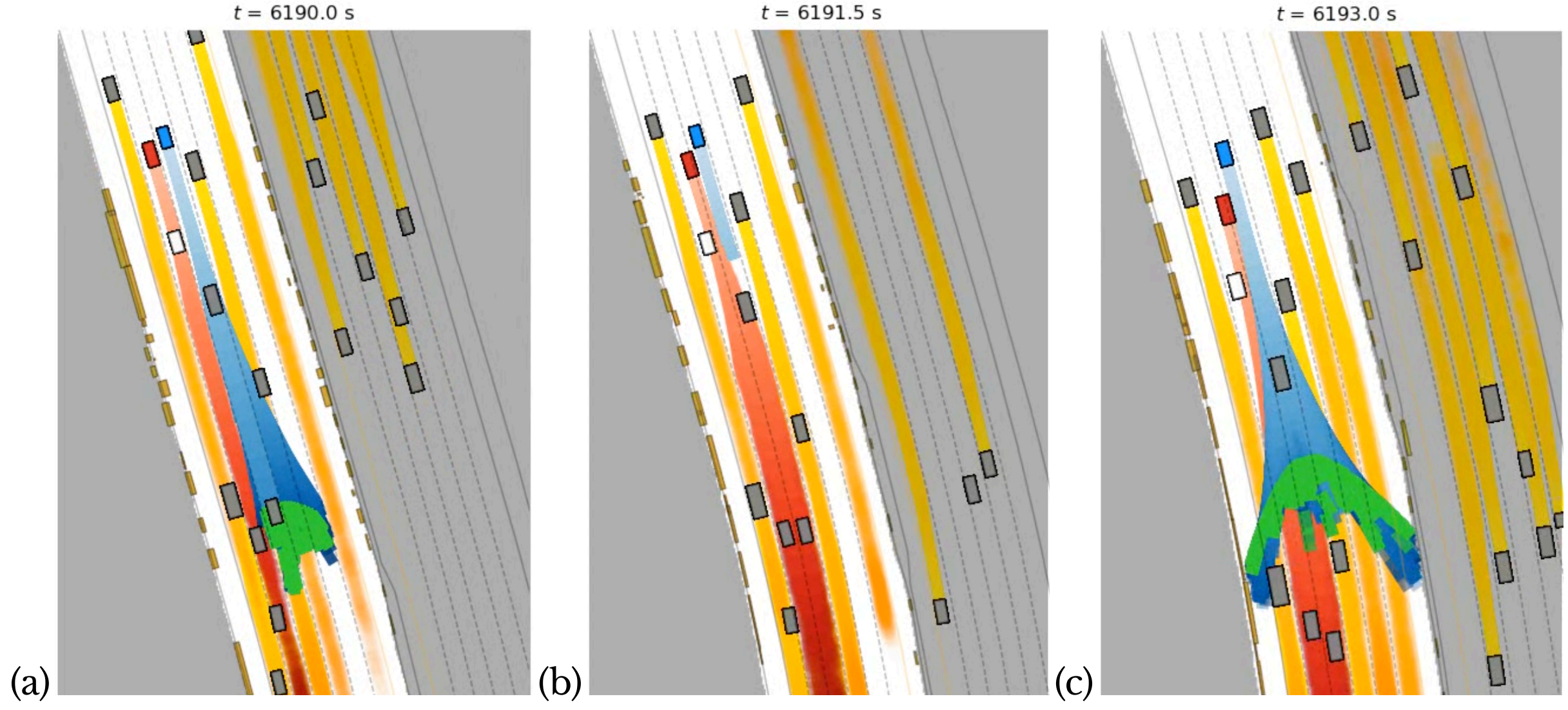

Figure 11. Visualization of reachable sets for the SDLI scenario. Agent initial states are indicated by outlined rectangles. The ego agent and its reachable sets are shown in blue, with the drivable area at the time horizon shown in green. ORUs are generally shown in gray, with reachable sets in orange, except for the cutting-in (initiator) ORU, which is shown in red. The initial state of the Waymo vehicle is shown in white. Permitted areas of the roadgraph for the ego are shown in white, and prohibited areas are shown in gray. At the beginning of the scenario (panel a), the ego is between an ORU on its left and the initiator ORU on its right. At the moment of the SDLI event (panel b), the initiator ORU that was on the ego's right has crossed into the ego's lane, cutting off its drivable area (note absence of green drivable area in panel b). After responding to the SDLI event by slowing (panel c), the ego vehicle has restored its drivable area.

## Straight crossing paths (SCP)

The FSM can be used to measure the availability of time gaps in traffic that afford movement. For example, if a vehicle is stopped at an intersection yielding to cross-traffic that has priority, the FSM can indicate when the yielding vehicle has an opportunity to pull out and merge onto the larger road. We applied the FSM to such a scenario that the Waymo Driver observed: in this scenario, a human driver waited at a stop sign for a suitable gap to appear in cross-traffic (see Figure 12).

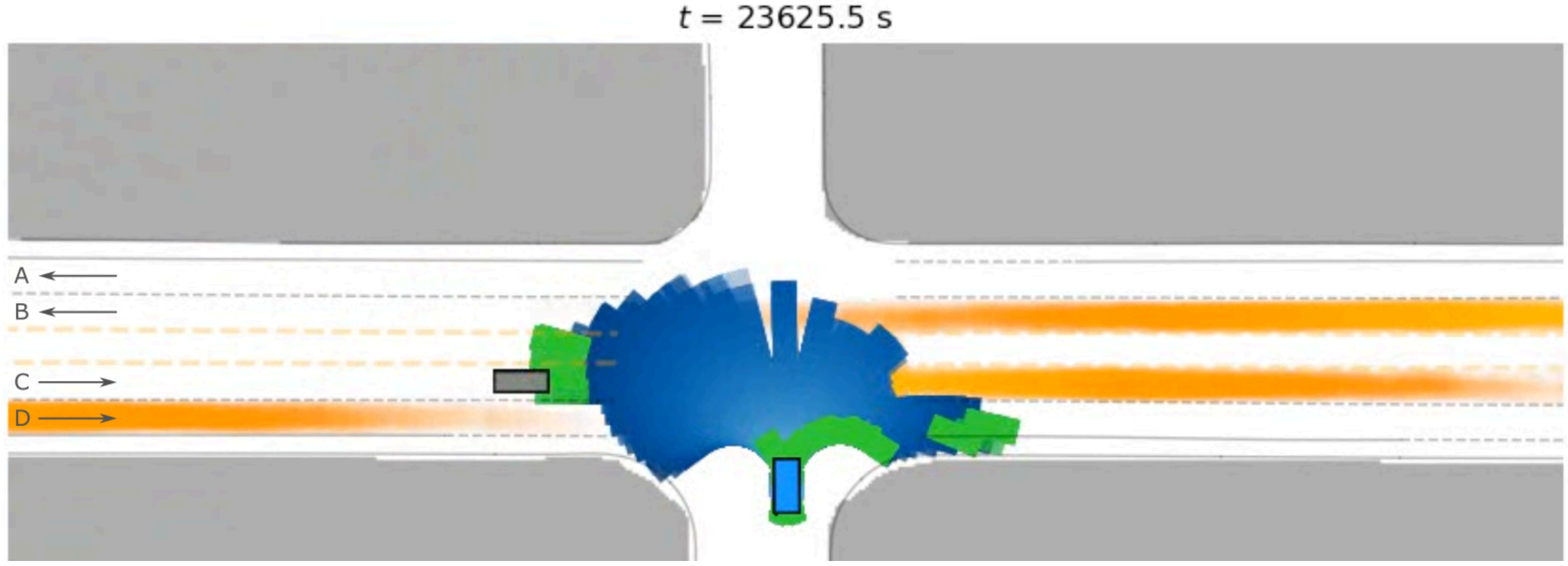

Figure 12. A human driver waits at a stop sign before merging onto a larger road. The larger road has two lanes of traffic from right to left (indicated on the left side using arrows and the letters A and B), a shared turn lane in the middle, and two lanes of traffic from left to right (arrows and the letters C and D). The ego vehicle's starting location is indicated using an outlined blue rectangle, with drivable areas shown in blue as future time $\tau$ increases from $0$ to $H$. ORUs are indicated using gray outlined rectangles, with reachable sets shown in orange. As in Figures 6 and 11, green areas show the ego's the ego's drivable area at the future time horizon, i.e., its set of conflict-free, kinematically reachable states at $\tau = H$. For example, green polygons in lane D, to the right of the ego, indicate that the ego has a conflict-free, kinematically reachable trajectory available for turning right and merging into the first lane. Similarly, the ego has an affordance to turn left into the shared turn lane, shown by green areas in the shared middle lane. In contrast, the lack of green polygons in lane C shows that the ego does not have a trajectory available for turning right and merging into the second lane: this trajectory overlaps with some reachable states from the ORU that is approaching the ego vehicle from the left in lane C. Similarly, there is no drivable area in lanes A or B, because the ego's trajectory leading into those lanes overlaps with the reachable sets of the vehicle approaching from the right in lane B.

The FSM indicates that the human driver has a conflict-free way to cross the first two lanes of traffic (coming from the left), but at that point would need to merge into the middle lane to avoid potential conflicts with traffic coming from the right.

The FSM thus reveals an affordance to cross the first two lanes, shown by the black polygons in Figure 11 that are to the left of the ego, in the center shared turn lane. Similarly, the FSM reveals that the ego does not have an affordance to cross all five lanes of the road at this moment, because there is no conflict-free reachable set for the ego that merges into either of the top two lanes on the road. Again, no additional parameter tuning was required from the base case kinematic assumptions.

## Opposite direction lateral incursion (ODLI)

In the scenarios discussed so far, the FSM has been used to operationalize the notion of maintaining a conflict-free "out" while driving. However, the FSM can also be used to analyze driving behavior during traffic conflicts.

We applied the FSM to a straight crossing paths scenario, recorded in a driving simulator, in which a human driver (the "subject vehicle" or SV) responded to an oncoming vehicle (the "principal other vehicle" or POV) that crossed suddenly into the SV's lane (Johnson et al., 2025; see figure 13). In this study, human drivers were often observed responding to the oncoming vehicle via a sequence of actions that extended over time: instead of a singular "response" to the conflict, people typically responded by slowing down first and then performing some other avoidance maneuver later.

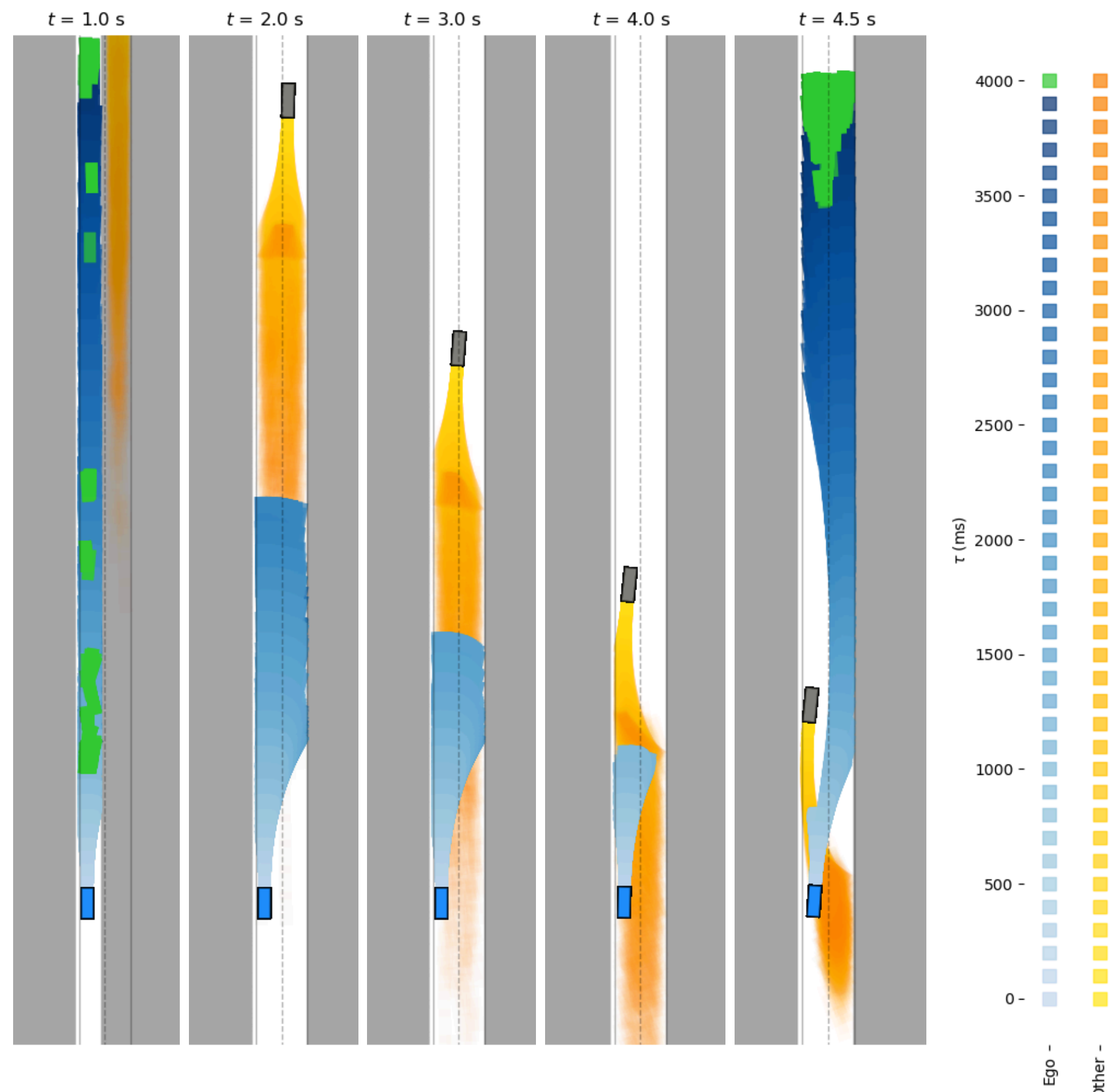

Figure 13. Visualization of reachable sets for an ODLI scenario in Johnson et al., (2025). The subject vehicle (SV, blue, at bottom) was responding to a principal other vehicle (POV, gray, with reachable sets in orange) that suddenly crossed into the SV's lane, going in the opposite direction. Note that this scenario took place in a left-hand driving context, as the driving simulator study was conducted in the UK, so the SV is traveling "up" and the POV is traveling "down."

Applying the FSM to the human responses indicated that the human drivers seemed to be sensitive to the affordances offered by the oncoming vehicle: in particular, human drivers tended to steer in response to the kinematic behavior of the POV, but only when the scenario provided an "out" for steering (Johnson et al., 2025). For example, if the POV crossed the SV's lane, participants had a tendency to steer into the POV's lane near the time when it became clear (based on the scenario kinematics) that the POV would not return to its own lane.

# Discussion

In this paper we presented the Field of Safe Motion, a reachability-based model for assessing observed driving behaviors. The FSM connects the psychologically motivated work of Gibson & Crooks (1938) with quantitative reachability models from the engineering literature (e.g., Althoff et al., 2008). We presented a summary of the model, its use of the roadgraph and ground plane raster, parametric assumptions used by the model, and potential output metrics that the model can provide. We then evaluated the model using several examples of naturalistic human driving behavior, including a tailgating scenario, a cut-in scenario, an intersection scenario, and an oncoming vehicle collision avoidance scenario. For the tailgating scenario we examined the effect of different kinematic assumptions on the model's outcomes, revealing that some assumptions can affect the model more than others. However, we obtained reasonable outcomes from all observed behaviors without needing to provide scenario specific parameter settings. Thus the FSM provides a well-motivated model for making quantitative assessments of driving behavior, across a range of scenarios, with minimal parameter tuning. Also, as noted above, the FSM is capable of identifying potential driving safety violations even in the absence of a conflict or collision.

Like other reachability approaches, the FSM can be expensive to compute, and the assumptions used by the model need to strike a balance between optimism and pessimism. However, we have incorporated several tuning parameters that allow a practitioner to trade off computational complexity for model fidelity, including, for example, the resolution of the ground plane raster, or the number of samples used in the εRANDUP algorithm (Lew et al., 2022). In general, any technique for speeding up reachability should be applicable for use in the FSM (e.g., Manzinger et al., 2020). Also, the FSM uses a simple representation of the world—the ground plane raster—that provides an interface for using non-kinematic models as needed, such as generative machine-learned models for predicting future states of agents. Such additional models could be integrated at the level of entire scenarios, or even for individual agents in a given scenario. Of course, integrating an opaque prediction model would limit interpretability, but this may be a worthwhile tradeoff for some applications.

The FSM performs kinematic rollouts in an open loop manner; that is, agents do not adjust their kinematic behaviors in response to hypothetical future states of the world. Furthermore, non-ego agents do not respond to each other while rolling out reachable sets. This limits the specificity of the FSM, but it is motivated by the reachability approach, where kinematic rollouts need to cover all possible future states, rather than sequences of action-reaction states. Implementing closed loop control would increase both (a) conceptual complexity, by adding the need for an action policy for each agent, and (b) runtime demands of the model, for computing actions based on other agents' potential future rollouts. In addition, adding closed-loop feedback control eliminates easy parallel computation. However, some existing models, such as the Safety Force Field (Nistér et al., 2019), do operate over agent policies and can provide closed-loop agent evaluations, although these models present different sets of tradeoffs than the FSM. Finally, some driving scenarios (e.g., oncoming collisions) might be more natural to model using closed-loop controllers than others.

Some safety-related models rely on so-called surrogate measures to identify unsafe behaviors. For example, time-to-collision (TTC) is widely used as a proxy for anticipating unsafe outcomes (c.f. Mullakkal-Babu et al., 2020; Lewis-Evans, 2012). However, TTC is difficult to generalize to curved trajectories or roadgraphs with multiple intersecting lanes. Reachability analysis can be used to generalize TTC to arbitrary trajectory interactions; as described earlier, reachable sets conceptually represent all kinematically possible trajectories for road users. As discussed, the FSM presents a natural way to represent a generalization of TTC that accounts for all possible "outs" in a given configuration of agents.

## Bounded uncertainty

The FSM tempers the pessimism of standard kinematic models with information derived from the roadgraph (e.g., lane assignments), which allows the model to take into account reasonably foreseeable behaviors of road users near the ego. This process creates well-defined and easily interpretable limits on the reachable sets of road users, thus setting bounds on the uncertainty of their kinematic states in the near future. Within the reachable set, a road user's particular kinematic state at a given point in the future is treated by the FSM as being uniformly unknown, rather than having an associated likelihood, which would introduce additional complexity and make the model more difficult to interpret (c.f., Broadhurst et al., 2005; Althoff et al., 2008; Lefèvre et al., 2014).

Uncertainty is not typically addressed by existing driving safety models. For example, uncertainty is not directly addressed in Responsibility-Sensitive Safety (Shalev-Shwartz et al., 2017) or the Safety Force Field (Nistér et al., 2019), both of which assume access to veridical kinematic information, and do not incorporate counterfactual futures. Other models do attempt to represent uncertainty, but largely this happens via probabilistic representations (e.g., He et al., 2022, Mullakkal-Babu et al., 2020). While probabilistic representations are a natural fit for representing different aspects of distributional futures, bringing probability into the picture when attempting to reason about future road occupancy tends to shift the discussion to a debate about "risk" that is difficult to ground empirically, especially for rare, long-tail, events. We find that the binary distinction of *bounded uncertainty* provided by the FSM—namely, "is it possible for this location to be reasonably foreseeably occupied by this agent at this time"—provides more clarity and interpretability when evaluating driving behavior.

## Assumptions

The FSM relies on a set of assumptions that mostly represent either (a) the limits of various kinematic variables in the model or (b) tradeoffs between model fidelity and runtime effort. In either case, parameters in the FSM have a clear interpretation and play a predictable role in the model. This stands in contrast with machine learning models, where model parameters do not affect the outcome in a clearly interpretable way. In the FSM, explaining how an agent is able (or unable) to achieve a particular state at some point in the future is a matter of understanding the physics equations that govern its motion, in combination with the constraints provided by the kinematic limits and the roadgraph. This interpretability is useful from a safety perspective since it provides some degree of explainability for an agent's behavior. Interpretability is also useful for engineers examining the behavior being assessed, to determine where and how to solve behavior problems.

The FSM, like any kinematic reachability model, relies heavily on assumptions about the kinematic capabilities of road users and their vehicles. Therefore the mechanism for setting these limits becomes centrally important to the model and its resulting measurements.

Many kinematic parameters are limited in an absolute sense by the physical characteristics of the vehicles in question. For example, the maximum braking capability of a typical passenger vehicle, with standard tires on dry road surfaces, can be estimated through modeling and/or tested empirically. However, assuming maximum possible braking or swerving capabilities may not be desired because road users might only rarely utilize the full range of kinematic capabilities.

Rather, assumptions in safety-related models are better thought of as depending on societal consensus about the *reasonable foreseeableness* of the assumed kinematic limits. For example, the reasonably foreseeable maximum braking or steering capability of a passenger vehicle is constrained by the known physical handling capacities of that vehicle, but can be further limited by considerations on what states should be considered reasonably foreseeable states, also taking societal normative expectations into account (e.g., the assumption that other vehicles will remain in their lane unless indicating otherwise).

The reasonable foreseeableness of kinematic assumptions in FSM may also be informed by empirical observation of the driving behaviors shown in real-world driving. For example, the assumed maximum deceleration threshold for a passenger car in the FSM could be set to some quantile of the braking deceleration values displayed by real drivers in a given scenario. This can be further fine-tuned by measuring the distributions of kinematic parameters in ODD-specific environments (e.g., the distribution of braking strengths on the freeway might differ from those on surface streets). This approach is advantageous because (a) it is relatively straightforward to measure observed kinematics, and (b) using limits derived from observed distributions provides clear bounds on the amount of kinematic behaviors that are *not* incorporated into the model. For example, setting the braking threshold to the 99th percentile of observed values in a given scenario indicates that the FSM with this assumption can be argued to include 99 percent of potential future rollouts in the real world. On the other hand, basing assumptions on observed distributions could lead to erroneous conclusions if the observed distribution does not match the scenario being evaluated—for example, a collection of observed driving behavior might not include rare events such as a loss of control.

IEEE 2846 (IEEE, 2022) provides a framework for formally defining such assumptions. As discussed above, the clear interpretability of the kinematic parameters in the FSM makes this sort of process relatively straightforward (as opposed to, say, the difficulty of setting the parameters of a machine-learning driving model via a consensus process).

## Extensions and future work

We foresee several extensions to the model that can be incorporated in future work. On a practical level, the FSM could incorporate more precisely tuned parameter limits by combining information from the roadgraph as well as from kinematic states to represent a broader range of normative expectations. For example, it is generally reasonable to assume that road users will come to a stop at traffic signals, so the FSM could set kinematic parameter limits or restrict roadgraph availability based on this information.

The FSM could also be extended to incorporate collision severity modeling (Kusano & Victor, 2022), to indicate whether a driver's behavior has exposed it to the potential for high-severity collisions. Such a feature would allow the FSM to differentiate between potentially unsafe driving behaviors with mild consequences versus those with severe consequences had a crash occurred.